\documentclass[a4paper,conference]{IEEEtran}

\usepackage{amsmath}

\usepackage{graphicx}
\usepackage{amsmath,amssymb}
\usepackage{color}
\usepackage{subfigure}
\usepackage{tabularx}
\usepackage{rotating}
\usepackage{diagbox}
\makeatletter
\renewcommand{\@thesubfigure}{\hskip\subfiglabelskip}
\makeatother
\usepackage{multirow}

\begin{document}

\title{Spatial-Aware GAN for Unsupervised Person Re-identification}

\author{
\IEEEauthorblockN{Changgong Zhang}
\IEEEauthorblockA{Artificial Intelligence Center\\
Alibaba DAMO Academy\\
China, 311121\\
Email: changgong.zhang@alibaba-inc.com}
\and
\IEEEauthorblockN{Fangneng Zhan}
\IEEEauthorblockA{School of Computer Science and Engineering\\
Nanyang Technological University\\
Singapore, 639798\\
Email: fnzhan@ntu.edu.sg}
}

\maketitle

\begin{abstract}
The recent person re-identification research has achieved great success by learning from a large number of labeled person images. On the other hand, the learned models often experience significant performance drops when applied to images collected in a different environment. Unsupervised domain adaptation (UDA) has been investigated to mitigate this constraint, but most existing systems adapt images at pixel level only and ignore obvious discrepancies at spatial level. This paper presents an innovative UDA-based person re-identification network that is capable of adapting images at both spatial and pixel levels simultaneously. 
A novel disentangled cycle-consistency loss is designed which guides the learning of spatial-level and pixel-level adaptation in a collaborative manner. In addition, a novel multi-modal mechanism is incorporated which is capable of generating images of different geometry views and augmenting training images effectively. Extensive experiments over a number of public datasets show that the proposed UDA network achieves superior person re-identification performance as compared with the state-of-the-art.
\end{abstract}

\IEEEpeerreviewmaketitle

\section{Introduction}
Person Re-Identification (ReID) aims at retrieving images of the same person from an image set collected with different cameras. It has been attracting increasing interest in both academia and industry in recent years thanks to its importance in surveillance and public security. Existing person ReID systems trained with a large number of labeled images have achieved very high accuracy, but they usually experience a dramatic performance drop while applied to images collected in a different environment due to the domain shift and bias. Similar to many other tasks, this has become a critical issue for person ReID which needs to handle images collected under different lighting, camera parameters, viewpoints, etc. for deployment in various different environments.

\begin{figure}
\centering
\includegraphics[width=0.95\linewidth]{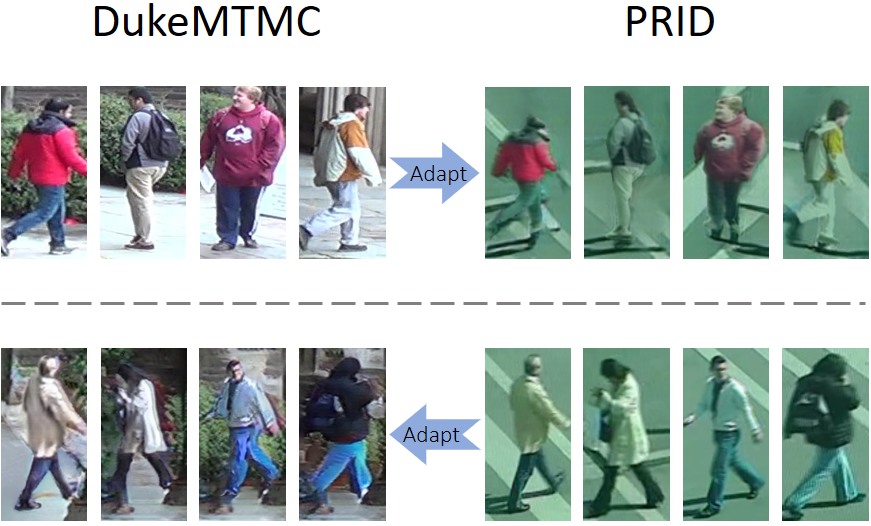}
\caption{Discrepancy exists at both pixel level (in image colors, image styles, etc.) and spatial level (in viewpoint, contextual structures, etc.) across domains. Our proposed Spatial-Aware GAN can adapt images in both spaces simultaneously (DukeMTMC and PRID are two public person ReID datasets that were collected under different lighting and viewpoints).}
\label{fig:intro}
\end{figure}

One typical strategy to tackle domain shift and bias is Unsupervised Domain Adaptation (UDA) which aims to transfer the learned knowledge from a labeled source domain to an unlabeled target domain. With the advances of Deep Neural Networks (DNNs) and Generative Adversarial Networks (GANs) \cite{goodfellow2014gan}, UDA has been applied to various image-to-image translation problems \cite{zhu2017cyclegan} as well as person ReID \cite{liao2018ptgan,deng2018spgan}. On the other hand, the UDA-based person ReID is still facing various constraints and its performance lags far behind fully supervised models. One major reason is that existing systems focus largely on pixel-level adaptation but ignore spatial discrepancies across domains. This can be seen clearly in Fig. \ref{fig:intro} where images in the source and target domains differ not only in colors and styles but also in viewpoints and contextual structures. The ignorance of geometric discrepancies directly leads to unfilled gaps at spatial level which further impairs the usefulness of the adapted images especially when such geometric discrepancy is large. At the other end, such geometric discrepancy acting as certain interference also affects the adaptation of image colors and styles at pixel level.

In this work, we propose an innovative Spatial-Aware Generative Adversarial Network (SA-GAN) to address the challenge of simultaneous image adaptation in both spatial and pixel levels. The proposed SA-GAN embeds spatial transformer modules into a cycle structure and achieves collaborative learning training at both spatial and pixel levels concurrently. The spatial transformer module incorporates a novel multi-modal mapping strategy for generating multiple images of different viewpoints, as well as a spatial discriminator for improving its stability and convergence during training. In addition, a novel disentangled cycle consistency loss is designed for the concurrent and smooth training at spatial and pixel levels in a collaborative manner. 

The contributions of this work are threefold. First, it designs an innovative UDA-based person ReID network that is capable of adapting images at both spatial and pixel levels simultaneously. Second, it designs a multi-modal mapping mechanism with a spatial discriminator that performs adversarial training at spatial level and is capable of mapping each source-domain image to multiple images with target-domain geometric characteristics. Third, it designs a novel disentangled cycle loss that balances the cycle-consistency for optimal training at spatial and pixel levels concurrently.

\section{Related Works}

\subsection{Person Re-identification}
Quite a number of person ReID systems have been reported in recent years which design different networks and architectures for optimal person ReID \cite{yi2014,wu2016}. 
\cite{yi2014} proposes a “Siamese” network for joint learning of color feature, texture feature and metric. 
\cite{wu2016} proposes an end-to-end network for simultaneous learning of high-level features and a corresponding similarity metric. 
% \cite{varior2016} presents a novel Siamese Long Short-Term Memory (LSTM) architecture to process image regions sequentially to enhance the discriminative capability of local feature representation. 
Some approaches position person ReID as a classification problem with the availability of person labels \cite{zheng2016,Wu2018,zheng2017}. 
\cite{zheng2016} proposes the ID-discriminative embedding (IDE) to train the re-ID model as classification task.
\cite{swu2016} presents a Feature Fusion Net (FFN) for pedestrian image representation by utilizing color histogram features and texture features. 
\cite{sun2017} adopts Singular Vector Decomposition (SVD) and proposes to optimize the deep representation learning with the restraint and relaxation iteration (RRI) training scheme. 
\cite{mcLaughlin2015} introduced a novel data augmentation method for re-identification based on changing the image background. 
\cite{zhong2017} introduces Random Erasing as a new data augmentation method for training the convolutional neural network. 
\cite{zhu2017} proposes a Pseudo Positive Regularization (PPR) method to enrich the diversity of the training data. 
Though the aforementioned networks achieve very high person ReID accuracy, they require a large amount of labeled images and thus have poor scalability while facing various new environments and domains.

In recent years, unsupervised domain adaptation (UDA) has been studied to mitigate the constraints of image labeling and annotation. \cite{zheng2017} exploits GANs to generate unlabeled samples and achieves improved person ReID performance. \cite{zhong2018camstyle} introduces the camera style (CamStyle) as a data augmentation approach to smooth the camera style disparity. \cite{liao2018ptgan} proposes a Person Transfer GAN (PTGAN) to bridge the domain gap. \cite{deng2018spgan} proposes a similarity-preserving GAN (SPGAN) that preserves the self-similarity in translation and dissimilarity across the translated domains. 
\cite{li2018} performs unsupervised domain adaptation which leverages information across datasets and derives domain-invariant features for Re-ID purposes.
The aforementioned UDA-based systems work largely at pixel level but ignore spatial discrepancy across domains. The proposed SA-GAN instead adapts images in both spaces simultaneously which achieves superior person ReID performance, more details to be described in the ensuing Sections.

\subsection{Domain Adaptation}
Domain adaptation was introduced to minimize the discrepancy across domains so as to obtain domain-invariant features \cite{saenko2010,torralba2011}. One typical approach is to minimize the distance in the feature space between the source and target domains. \cite{sun_2016_2} performs the feature adaptation by minimize the correlation distance and \cite{sun2016} extended it to deep architectures. \cite{long2017} instead strive to minimize the Maximum Mean Discrepancies (MMD) and Joint MMD distance across domains. \cite{bousmalis2016} explicitly models domain-specific features to encourage networks to learn domain-invariant features. \cite{tzeng2017} further improve the feature adaptation by introducing the adversarial objective.

Another typical approach is to minimizing pixel-level distances by directly converting the style of the source domain to the style of the target domain. This is mostly accomplished by using GANs which have achieved great success in 
image generation \cite{radford2016dcgan,arjovsky2017wgan}, 
image composition \cite{lin2018stgan,zhan2019sfgan,zhan2019acgan,zhan2018verisimilar,zhan2019scene,zhan2020towards,zhan2019esir,zhan2020emlight,zhan2020aicnet} and
image-to-image translation \cite{zhu2017cyclegan,isola2017pixel,zhan2019gadan}. For domain adaptation, \cite{taigman2017} presents a domain transfer network that transforms a source image to a target image by enforcing consistency in the embedding space. 
\cite{bousmalis2017} uses a content similarity loss to ensure the similarity between the generated target image and the original source image. \cite{donahue2017} instead learn the transformations in the pixel and latent spaces simultaneously. More recently, CycleGAN \cite{zhu2017cyclegan} 
and similar methods \cite{yi2017dualgan,kim2017,hoffman2018cycada} 
produces compelling image translation results at high resolution by using the cycle-consistency loss. 

\begin{figure*}[t]
\centering
\includegraphics[width=0.995\linewidth, height=0.475\linewidth]{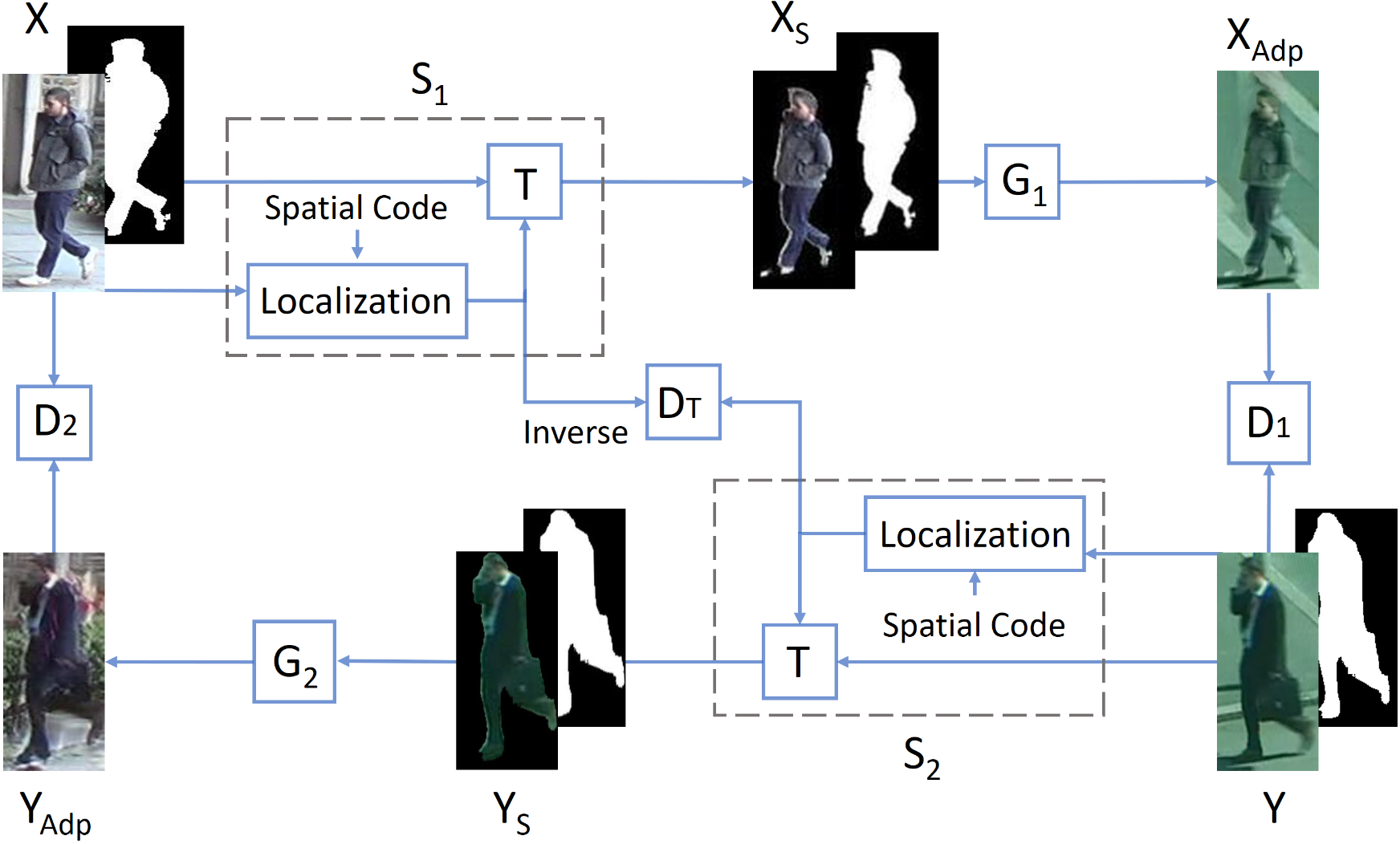}
\caption{The structure of the proposed Spatial-Aware GAN (SA-GAN): The parts within the dashed lines are two Spatial Transformer Modules $S_1$ and $S_2$ each of which consists of a spatial transformation $T$ and a parameter localization network $Localization$. $G_1$, $G_2$, $D_{1}$, $D_{2}$ and $D_{T}$ denote the generators and discriminators, respectively. $X$ and $Y$ denote two image domains (the binary masks are determined by U-Net), where ($X_s$, $Y_s$) and ($X_{adp}$, $Y_{adp}$) denote the two-domain images after the proposed spatial-level and further pixel-level adaptation, respectively. }%For clarity, cycle loss and identity loss are not included.
\label{fig:stru}
\end{figure*}
% There is not 'Localization' in the graph. Change Estimtation1 and Estimation2 to 'Localization'? - Shijian
% change

\section{Proposed Method}
We design an innovative spatial-aware generative adversarial network (SA-GAN) for multi-modal domain adaptation at spatial and pixel levels simultaneously. Fig. \ref{fig:stru} shows the detailed network structure and training strategy as to be described in the following subsections.

\subsection{Network Structure}
The proposed SA-GAN consists of spatial transformer modules (STMs) $S_1$ and $S_2$, generators $G_{1}$ and $G_{2}$, and discriminators $D_{1}$, $D_{2}$, and $D_{T}$ as illustrated in Fig. \ref{fig:stru}. It is designed in a cyclic structure, aiming to achieve the adaptation from domain $X$ (source) to domain $Y$ (target) at both pixel and spatial levels simultaneously.

As illustrated in Fig. \ref{fig:stru}, the space-level adaption is largely achieved by the STMs ($S_1$ and $S_2$) and the spatial discriminator $D_{T}$. The STMs apply spatial transformations to  source-domain images to adapt them to have similar geometries with  target-domain images. The transformed images $X_{S}$ are then adapted at pixel level to have similar appearance with the target-domain images, marked as $X_{Adp}$. The pixel-level adaption is achieved by the $G_{1}$, $G_{2}$, $D{1}$, and $D_{2}$, where $D_{1}$ discriminates images $X_{Adp}$  and $Y$, while $D_{2}$ discriminates images $Y_{Adp}$ (images adapted from domain Y and domain X) and $X$. The two discriminators work together to improve the learning of generators $G_{1}$ and $G_{2}$ at pixel level. 

We also introduce a Siamese loss through Siamese Network (SiaNet) to preserve the semantic of the adapted images. The positive pairs are $X \& X_{s}$, $X \& X_{Adp}$, and the negative pairs are $X \& Y_{s}$, $X \& Y$. The input of the SiaNet contains a pair of images, where the objective is to minimize the distance between the positive pairs and enlarge the distance between negative pairs.

A binary human mask is concatenated with the origin image as input of the model, it is also used for a weight of the identity loss (as 3.3(5)). We pre-trained a U-net \cite{unet} off-line on Look into Person (LIP) dataset \cite{2017gonglip} for the function of the binary human mask. It is then applied for inference to predict the foreground of person ReID images.

% It may not be suitable to put the paragraph above in this subsection. In addition, the content is related to Fig. 3 but there is no corresponding text description to link up with Fig. 3. - Shijian

It should be noted that the image adaptation does not perform well by direct concatenation of spatial-level adaptation GANs (e.g. ST-GAN \cite{lin2018stgan}) and pixel-level adaptation GANs (e.g. CycleGAN \cite{zhu2017cyclegan}). The major reason is that images in source and targets domains have discrepancies at both spatial and pixel levels. By directly concatenating these two types of GANs, the two kinds of discrepancies will act as certain negative interference to affect each other. Our SA-GAN instead coordinates the learning at spatial and pixel levels concurrently, where better adaptation at spatial level drives the model to learn better adaptation at pixel level and vice versa. The cooperative and concurrent learning between spatial and pixel levels also drives the model to converge stably and efficiently.

\begin{figure*}[t]
\centering
\includegraphics[width=0.95\linewidth, height=0.3\linewidth]{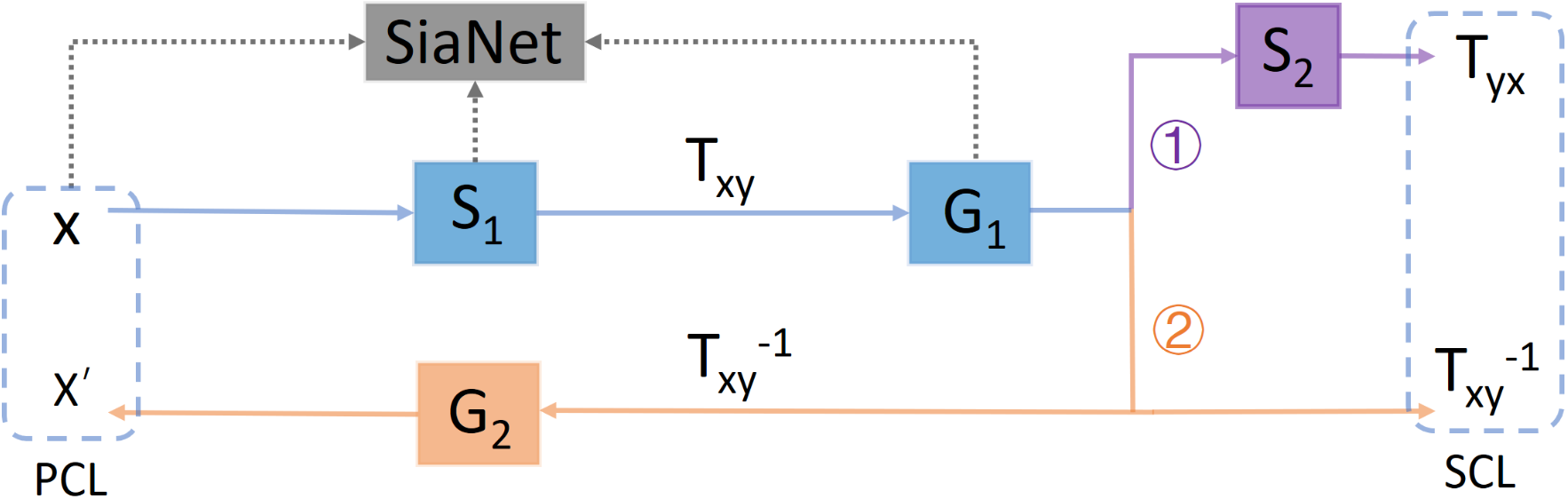}
\caption{Cycle-consistency loss decomposition. The part in purple denotes Path 1 in which $S_{2}$ recover the image at spatial level. The part in orange denotes Path 2 in which the inverse matrix of $T_{xy}$ is used to recover the image. The part in gray denotes the Siamese loss.}
\label{fig:loss}
\end{figure*}

\subsection{Spatial Level Adaptation}
The proposed SA-GAN incorporates a novel multi-modal strategy for spatial-level adaptation. Specifically, the STM $S_1$ (or $S_2$) consists of a localization network and a transformer $T$. In $S_1$ (or $S_2$) for mapping $X \rightarrow Y$ (or $Y \rightarrow X$) at spatial level, a spatial code sampled from normal distribution is concatenated with $X$ (or $Y$) as the input of the localization network to predict the spatial transformation parameters which can be affine, homography or thin plate spline \cite{tps}. By sampling multiple spatial codes, the estimation network predicts multiple different spatial transformations which produces multiple adapted images with similar geometric characteristics of the target domain.

$D{1}$ and $D_{2}$ drive the learning process at spatial and pixel levels, but they are sometimes insufficient for a optimal learning and may even lead to a convergence problem due to the high flexibility of the spatial transformer module and the entangled learning in both spaces concurrently. Hence we introduce a spatial discriminator $D_{T}$ into SA-GAN for optimal learning at spatial level. Instead of distinguishing images, $D_{T}$ discriminates the predicted spatial transformation $X \rightarrow Y$ and the inverse spatial transformation $Y \rightarrow X$. As $X \rightarrow Y$ and $Y \rightarrow X$ are mapping in reverse direction, the inverse of the spatial transformation $Y \rightarrow X$ should be similar to the transformation $X \rightarrow Y$. By distinguishing the two transformation matrixes, $D_{T}$ imposes an extra constraint on  learning of the spatial transform modules $S_1$ and $S_2$ which greatly improves the effectiveness and stability of the learning process at spatial level.

\subsection{Training Loss}
The proposed SA-GAN is designed in a cycle structure, where ($S_{1}$, $G_{1}$) and ($S_{2}$, $G_{2}$) deal with the mappings $X \rightarrow Y$ and $Y \rightarrow X$, respectively. 
We adopt the Wasserstein GAN \cite{arjovsky2017wgan} objective to learn the mapping $X \rightarrow Y$ and the loss functions are formulated as follows:
\begin{equation}
\begin{split}
& L_{D_{adv}} = E_{x \sim X }[D_{1}(G_1(S_{1}(x)] - E_{y \sim Y}[D_{1}(y)] \\
& + E_{x \sim X }[D_{T}(T_{xy})] - E_{y \sim Y}[D_{T}((T_{yx}){-1})] \\
& L_{G,S_{adv}} = -E_{x \sim X}[D_{Y}(G_1(S_{1}(x)] - E_{x \sim X}[D_{T}(T_{xy})] \\
\end{split}
\end{equation}
where $T_{xy}$ and $T_{yx}$ denote the transformations $X \rightarrow Y$ and $Y \rightarrow X$, respectively. 

It is a challenging task to predict the spatial transformations accurately in the cycle process, as a very small error in the predicted spatial transformation parameters could lead to a large overall cycle-consistency loss which further overwhelms the pixel-level cycle-consistency loss. To solve this problem, we designed a disentangled cycle-consistency loss for a smooth and stable image adaptation at both spatial and pixel levels. Specifically, we disentangle the overall cycle consistency loss into a spatial-level cycle-consistency loss (SCL) and a pixel-level cycle-consistency loss (PCL), and use them for image adaptation at spatial and pixel levels, respectively.

As shown in Fig. \ref{fig:loss}, the prediction of input image $x$ by $S_{1}$ is first adapted at spatial level by spatial transformation $T_{xy}$ and  then at pixel level by $G_{1}$. There are two paths for the calculation of cycle-consistency loss. In the first path, $S_{2}$ predicts a spatial transformation $T_{yx}$ (using the same spatial code as $S_{1}$) to recover the image at spatial level as highlighted by purple color. In the second path, the output of $G_{1}$ is transformed by an inverse spatial transformation $(T_{xy})^{-1}$ and then adapted by $G_{2}$ to obtain the image $x^{'}$ (as highlighted in orange color) which is perfectly recovered in spatial level.

\renewcommand\arraystretch{1.8}
\begin{table*}[t]
\caption{Unsupervised person ReID performance by the proposed SA-GAN and state-of-the-art methods: The source dataset is Market-1501 and the target datasets are DukeMTMC on the left and DukeMTMC on the right. `Original' refer to the baseline model that is trained on the source dataset and evaluated on the target datasets without adaptation.}
\centering
\begin{tabular}{|l|p{1.5cm}<{\centering} | p{1.5cm}<{\centering} | p{1.5cm}<{\centering} | p{1.5cm}<{\centering} | p{1.5cm}<{\centering} | p{1.5cm}<{\centering} |}\hline
\multirow{2}{*}{Methods} & \multicolumn{3}{c|}{DuekMTMC to Market-1501} & \multicolumn{3}{c|}{Market-1501 to DukeMTMC} \\
\cline{2-7} 
                                 & R-1  & R-10 & mAP  & R-1  & R-10 & mAP  \\\hline
Original                         & 36.8  & 59.4 & 14.3 & 25.4 & 46.2 & 11.2 \\
CycleGAN \cite{zhu2017cyclegan}  & 40.5 & 62.7 & 16.5  & 32.4 & 53.4 & 16.1 \\
CycleGAN* \cite{zhu2017cyclegan} & 42.9 & 64.2 & 16.7 & 33.7 & 53.9 & 16.8 \\
TJ-AIDL \cite{wang2018tj-aidl}   & 58.2 & 81.1 & 26.5 & 44.3 & 65.0 & 23.0 \\
PT-GAN \cite{liao2018ptgan}      & 38.6 & 66.1 & - 
& 40.2 &  -   & 21.8 \\
SPGAN \cite{deng2018spgan}       & 51.5 & 76.8 & 22.8 & 41.1 & 63.0 & 22.3 \\
MMFA \cite{lin2018mmfa}          & 56.7 & -    &  -   & 45.3 & 66.3 & 24.7 \\
Camstyle \cite{zhong2018camstyle}& 58.8 & 84.3 &  27.4 & 48.4 & \textbf{68.9} & 25.1 \\
HHL \cite{zhong2018generalizing} & 62.2 & 84.0 &  \textbf{31.4} & 46.9 & 66.7 & 27.2 \\
\hline
SA-GAN                           & 59.3 & 82.7 & 27.9 & 46.8 & 67.1 & 26.9 \\
SA-GAN [M=10]                    & \textbf{63.1} & \textbf{84.5} & 30.7 & \textbf{49.5} & 68.2 & \textbf{27.9} \\\hline
\end{tabular}
\label{tab:market}
\end{table*}

\renewcommand\arraystretch{1.8}
\begin{table}[t]
\caption{Unsupervised person ReID performance by the proposed SA-GAN and state-of-the-art methods: The source dataset is Market-1501 and the target datasets are PRID. `Original' refer to the baseline model that is trained on the source dataset and evaluated on the target datasets without adaptation.}
\centering
\begin{tabular}{|l|p{1.2cm}<{\centering} | p{1.2cm}<{\centering} | p{1.2cm}<{\centering} |}\hline
\multirow{2}{*}{Methods} & \multicolumn{3}{c|}{Market-1501 to PRID} \\
\cline{2-4} 
                                 & R-1  & R-10 & mAP   \\\hline
Original                         & 7.8  & 20.3 & 3.1  \\
CycleGAN \cite{zhu2017cyclegan}  & 14.5 & 31.7 & 6.5 \\
CycleGAN* \cite{zhu2017cyclegan} & 14.9 & 32.2 & 6.7 \\
TJ-AIDL \cite{wang2018tj-aidl}   & 26.8 & -    &  -  \\
PT-GAN \cite{liao2018ptgan}      & 32.6 & 70.3 & 20.7 \\
% SPGAN \cite{deng2018spgan}       &  -   & -    &  -   \\
MMFA \cite{lin2018mmfa}          & 35.1 & -    &  - \\\hline
SA-GAN                           & 37.2 & 76.1 & 22.9 \\
SA-GAN [M=10]                    & \textbf{41.7} & \textbf{80.7} & \textbf{25.3}  \\\hline
\end{tabular}
\label{tab:prid}
\end{table}

As the spatial transformation is totally recovered in $x'$, there is only pixel-level difference between $x$ and $x'$. We therefore formulate the PCL of $x$ as follows:
\begin{equation}
\begin{split}
PCL_{X} = E_{x \sim X}[\left \| x' - x \right \|]
\end{split}
\end{equation}
For the spatial-level cycle-consistency, the $T_{yx}$ as predicted by $S_{2}$ should be similar to the inverse of $T_{xy}$. We therefore formulate the SCL as follows:
\begin{equation}
\begin{split}
SCL_{X} = E_{x \sim X}[\left \| (T_{xy})^{-1} - T_{yx} \right \|]
\end{split}
\end{equation}
The overall cycle-consistency loss can thus be formulated as follows:
\begin{equation}
\begin{split}
L_{cyc} = SCL_{X} + \lambda_{pcl} PCL_{X}
\end{split}
\end{equation}
where $\lambda_{pcl}$ is the weights of $PCL_{X}$.

We also introduce a mask identity loss to ensure that the translated image preserves feature of the human region:
\begin{equation}
\begin{split}
L_{idt} = E_{x \sim X} [\left \| G_1(S_{1}(x)) * M - S_{1}(x) * M \right \|]
\end{split}
\end{equation}
where $M$ denotes the human mask. 

The Siamese loss is calculated through a Siamese Network (SiaNet) as shown in Fig. \ref{fig:loss}, and loss function can be defined as below: 
\begin{equation}
\begin{split}
L_{siam}(i, x_{1}, x_{2}) = (1-i)(max(0, m-d))^2+id^{2}
\end{split}
\end{equation}
where $i=0,1$ denotes the negative or positive pair, $d$ denotes the Euclidean distance between the two input vectors and $m$ is the margin that defines the separability in the embedding space.

The overall loss for $G_{1}$ and $S_{1}$ can thus be formulated as follows:
\begin{equation}
\begin{split}
L_{G_{1}, S_{1}} = L_{G,S_{adv}} + \lambda_{cyc} L_{cyc} + \lambda_{idt} L_{idt} + \lambda_{siam} L_{siam}
\end{split}
\end{equation}
where $\lambda_{cyc}$, $\lambda_{idt}$ and $\lambda_{siam}$ denote the weights of the cycle-consistency loss, identity loss and Siamese loss, respectively. Similar loss can be formulated for $G_{2}$ and $S_{2}$.

\section{Experiments}
\subsection{Datasets}
We experimented proposed network in three popular Person reID public datasets as below to prove effectiveness.

\textbf{Market-1501} \cite{zheng2015bow} contains 32,668 labeled images of 1,501 identities that were collected from 6 camera views.
The dataset is split into two fixed parts: 12,936 images from 751 identities for training and 19,732 images from 750 identities for testing. The training set has 17.2 images per identity on average. In testing, 3,368 hand-drawn images from 750 identities are used as queries to retrieve the matching persons in the database. Single-query evaluation is used.

\textbf{DukeMTMC} \cite{zheng2017_2} contains 1,812 identities and 36,411 bounding boxes. 16,522 bounding boxes of 702 identities are used for training. The rest identities are used for testing. DukeMTMC is also denoted as Duke for short.

\textbf{PRID} \cite{hirzer2011} contains 934 identities from two camera views A and B. There are 385 identities in View A and 749 identities in View B, but only 200 identities appear in both views. 
% We use the bounding boxes of the 200 identities (appearing in views A and B) as testing set.

\subsection{Implementation}
% As Market-1501 
We use the ID-discriminative Embedding (IDE) \cite{zheng2016} as the baseline model which is trained with the strategy as described in \cite{zheng2016}. ResNet-50 \cite{resnet} is used as the backbone, and we only change the output dimension of the last fully-connected layer to be the number of training identities in corresponding datasets. The SGD solver is used to train re-ID model with a batch size of 128. The learning rate starts with 0.01 and will be divided by 10 after 25 epochs with total epoch of 50. In testing, we extract the output of the 2,048-dim vectors of Pool-5 layer as image descriptor and use the Euclidean distance to compute the similarity between images.

As the spatial-level discrepancy is mainly introduced by different camera views, we adopt homography transformation in the spatial transformer module which can convert among images of different viewpoints. The images in DukeMTMC and Market-1501 are captured with multiple cameras including low and high perspectives. We separate their training images into two sets according to the camera perspective and achieve the adaptation of the two sets separately.

\begin{figure*}[t]
\centering
\includegraphics[width=0.99\linewidth]{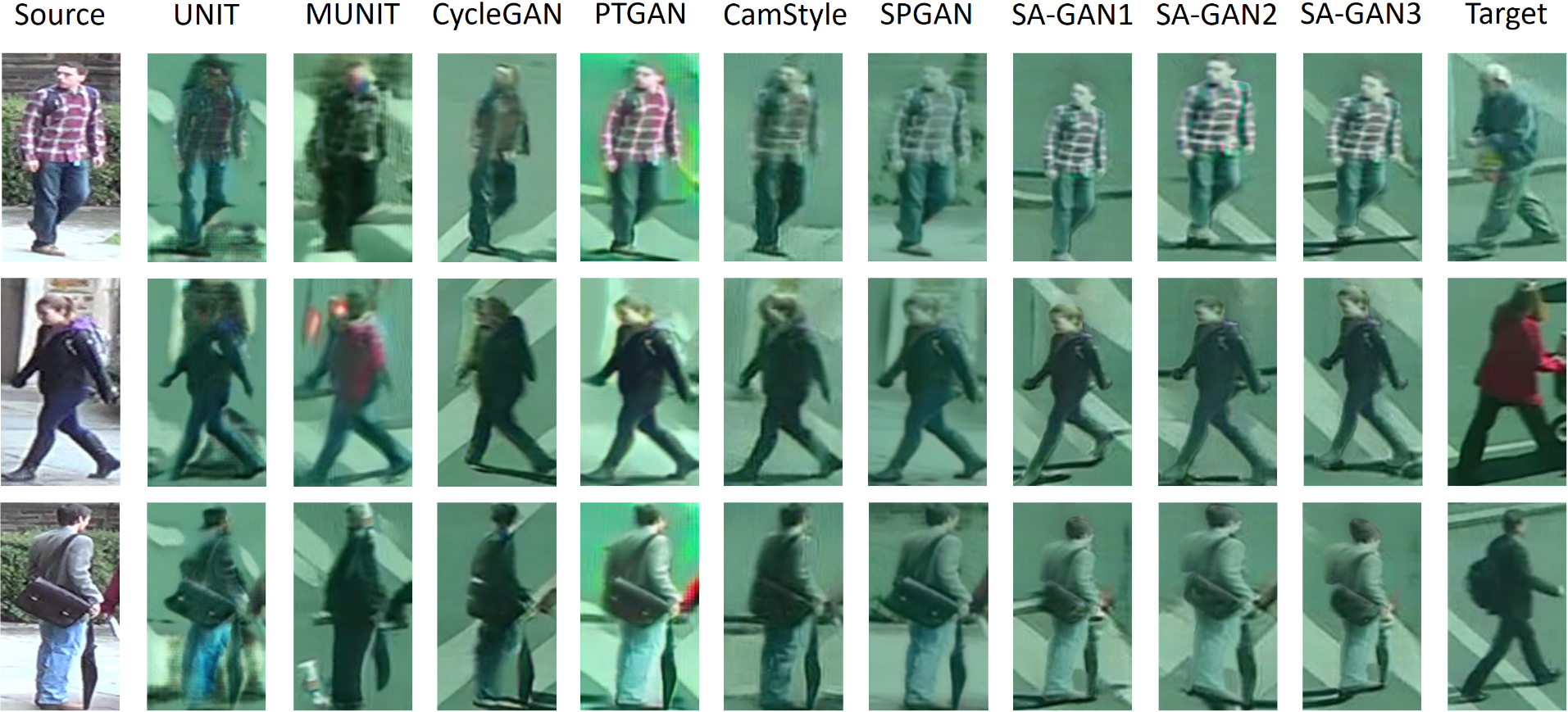}
\caption{Image adaptation from DukeMTMC (with a low viewpoint) to PRID CamB (with a high viewpoint) by SA-GAN and state-of-the-art methods: SA-GAN1, SA-GAN2, and SA-GAN3 refer to 3 adaptation (with 3 spatial codes) for each of the source images in the first column. The \textbf{Source} shows the images to be adapted by the listed methods and the \textbf{Target} shows images \textit{randomly sampled} from the target dataset (not paired one-to-one adaptation).}
\label{fig:comp}
\end{figure*}
% Better change 'origin' to Source. - Shijian
% changed

\subsection{Experimental Results}

\renewcommand\arraystretch{1.8}
\begin{table*}[t]
\caption{Ablation study of the proposed SA-GAN for adaptation from Duke-MTMC to PRID: PRID-CamA means using the CamA images as the query and CamB images as gallery, and vice versa. WS, WP, and WD denote `without spatial-level adaptation', `with pixel-level adaptation', and `without disentangled cycle loss', respectively. IDE \cite{zheng2016} is used as the baseline model.}
\centering % centering table
\begin{tabular}{|l| p{1.0cm}<{\centering} | p{1.1cm}<{\centering} | p{1.1cm}<{\centering} | p{1.1cm}<{\centering} | p{1.1cm}<{\centering} | p{1.1cm}<{\centering} | p{1.1cm}<{\centering} | p{1.0cm}<{\centering} |} % creating 10columns
\hline % inserting double-line
\multirow{2}{*}{Methods} & \multicolumn{4}{c|}{PRID-CamA} & \multicolumn{4}{c|}{PRID-CamB}\\
\cline{2-9}
                & R-1  & R-5  & R-10 & mAP  & R-1  & R-5  & R-10 & mAP \\\hline
\hline
Original        & 3.1  & 7.8  & 11.9 & 4.7  & 2.6  & 5.5  & 8.8  & 3.9  \\
SA-GAN (WP)     & 10.3 & 17.1 & 20.0 & 8.5  & 9.0  & 20.4 & 24.5 & 9.5  \\
SA-GAN (WD)     & 20.1 & 35.6 & 41.7 & 15.7 & 20.6 & 41.4 & 46.3 & 16.2 \\
SA-GAN (WS)     & 32.8 & 64.5 & 71.1 & 20.9 & 31.4 & 63.2 & 69.3 & 19.7 \\
SA-GAN [M=1]    & 34.4 & 66.7 & 73.3 & 21.1 & 32.7 & 65.4 & 71.7 & 20.3 \\
SA-GAN [M=10]   & \textbf{35.3} & \textbf{67.8} & \textbf{74.4} & \textbf{21.5} & \textbf{33.7} & \textbf{66.3} & \textbf{72.6} & \textbf{20.8} \\
\hline
% Model & 89.1 & 70.1 & 97.8 & 96.1 \\\hline
\end{tabular}
\label{tab:duke2prid}
\end{table*}

\textbf{DukeMTMC to Market-1501:}
To demonstrate that the proposed unsupervised person ReID method can work well under a general scenario, we perform another experiment for adaptation from \textit{Market-1501} to \textit{DukeMTMC}. Table \ref{tab:market} shows experimental results. The 'Original' refers to a baseline model that is trained over DukeMTMC and directly tested over \textit{Market-1501}. As Table \ref{tab:market} shows, the direct transfer can reach a R-1 accuracy of 36.8 \% and mAP of 14.3 \% respectively.
% , which is much higher than the adaptation from Market-1501 to PRID due to the much smaller discrepancy between Market-1501 and DukeMTMC.
Then we apply CycleGAN for pixel-level adaptation from \textit{DukeMTMC} to \textit{Market-1501}, and experiments show a 3.7\% improvement in R-1 accuracy.
We also benchmark our network with serveral state-of-the-art unsupervised methods including TJ-AIDL \cite{wang2018tj-aidl}, PTGAN \cite{liao2018ptgan}, SPGAN \cite{deng2018spgan} and MMFA \cite{lin2018mmfa}. As Table \ref{tab:market} shows, these unsupervised methods improve the person ReID performance greatly and achieve the best R-1 accuracy of 58.8 \% and mAP of 31.4 \%, respectively. 
The proposed SA-GAN (without multi-modal transformation) achieves a R-1 accuracy of 58.3 \% which is 0.5 \% lower than the state-of-the-art.
The improvement of 0.9\% can be observed when the proposed multi-modal transformation is included (M=10 spatial codes are used).
Note even the SA-GAN[M=10] has an lower mAp than HHL \cite{zhong2018generalizing}, as HHL adopts a strong baseline which reaches a mAP of 16.9\% while our baseline only reaches 14.3\%.

\textbf{Market-1501 to DukeMTMC:}
Table \ref{tab:market} shows experimental results for adaptation from \textit{Market-1501} to \textit{DukeMTMC}.
As Table \ref{tab:market} shows, the direct transfer can reach a R-1 accuracy of 25.4 \% and mAP of 11.2 \% respectively.
A clear 7\% improvement in R-1 accuracy is observed when applying CycleGAN for pixel-level adaptation from \textit{Market-1501} to \textit{DukeMTMC}.
Compared with the methods in the table, the proposed SA-GAN[M=10] achieves the best R-1 and mAP score, which is 1.1\% and 0.7\% higher than the state-of-the-art.

\textbf{Market-1501 to PRID:}
We first evaluate our proposed network for adaptation from dataset \textit{Market-1501} to dataset \textit{PRID}, as images in the two dataset contain clear discrepancy at both spatial and pixel levels. The evaluation protocol on PRID follows the same single-shot experiments as described in \cite{zhang2016}. Table \ref{tab:prid} shows experimental results. The `Original' in Table \ref{tab:prid} refers to a baseline model that is trained over \textit{Market-1501} and directly tested over \textit{PRID}. In addition, we also apply the CycleGAN \cite{zhu2017cyclegan} for image-to-image translation from \textit{Market-1501} to \textit{PRID}. As Table \ref{tab:prid} shows, the CycleGAN-translated images outperform the baseline model `Original' by 6.7 \% in R-1 accuracy and 3.4 \% in mAP, almost doubling the R-1 and mAP scores due to the significant pixel-level discrepancy between Market-1501 and PRID images. On the other hand, CycleGAN still scores a very low R-1 score of 7.8 \% because it tends to over-translate images and leads to the loss of person ID information. By including an identity loss into the CycleGAN to preserve more identify information as denoted by CycleGAN*, the R-1 and R-10 scores are improved by 0.4 \% and 0.5 \%, respectively.

We also benchmark our network with three state-of-the-art unsupervised person ReID networks including Transferable Joint Attribute-Identity Deep Learning (TJ-AIDL) \cite{wang2018tj-aidl}, PT-GAN \cite{liao2018ptgan} and Multi-task Mid-level Feature Alignment (MMFA) network \cite{lin2018mmfa}.
As Table \ref{tab:prid} shows, all three state-of-the-art networks clearly outperform the `Original' and CycleGAN. TJ-AIDL \cite{wang2018tj-aidl} improves the R-1 score by 19 \% by learning attribute-semantic and identity discriminative features simultaneously. PT-GAN introduces human masks to preserve the person ID information in adaptation and this helps improve R-1, R-10 and mAP by 24.8 \%, 50 \% and 17.6 \% respectively, as compared with CycleGAN. MMFA \cite{lin2018mmfa} reaches a R-1 score of 35.1 \% by assuming that the source and target datasets share the same set of mid-level semantic attributes. 

Our SA-GAN (without multi-modal transformation) outperforms the MMFA by 2.1\% in R-1 accuracy, largely due to its concurrently adaptation at both pixel and spatial levels. In addition, the R-1 accuracy is further improved to 39.7 \% when ten spatial codes are used in image adaptation as denoted by SA-GAN [M=10].

\textbf{Qualitative Results:}
Fig.~\ref{fig:comp} shows image adaptation results from DukeMTMC to PRID CamB by our proposed SA-GAN and several state-of-the-art methods, where SA-GAN1, SA-GAN2 and SA-GAN3 denote three adaptation results by SA-GAN with three random spatial codes. For the image adaptation from DukeMTMC to PRID CamB, there are not only pixel-level discrepancy but also spatial-level discrepancy because the source images are captured in a low viewpoint while the target images are captured in a high viewpoint. As Fig. \ref{fig:comp} shows, the proposed SA-GAN achieves image adaptation at both spatial and pixel levels simultaneously. Specifically, it is capable of transforming a source image to multiple target images of different spatial layout as illustrated in SA-GAN1, SA-GAN2 and SA-GAN3. Additionally, the adapted images have the same style as the target-domain images without losing person identity and the viewpoints are adapted properly as well. %spatial layout of them are also adapted to high viewpoints.
As a comparison, UNIT \cite{liu2017unit}, MUNIT \cite{huang2018munit} and CycleGAN \cite{zhu2017cyclegan} can only perform pixel-level adaptation, and the person ID information is impaired greatly. PTGAN \cite{liao2018ptgan}, SPGAN \cite{deng2018spgan} and Camstyle \cite{zhong2018camstyle} are capable of preserving certain person ID information, but their adaptation is still restricted within at pixel level and the adapted images are also quite different from the PRID images due to spatial-level discrepancies.

\begin{figure}
\centering
\includegraphics[width=0.95\linewidth]{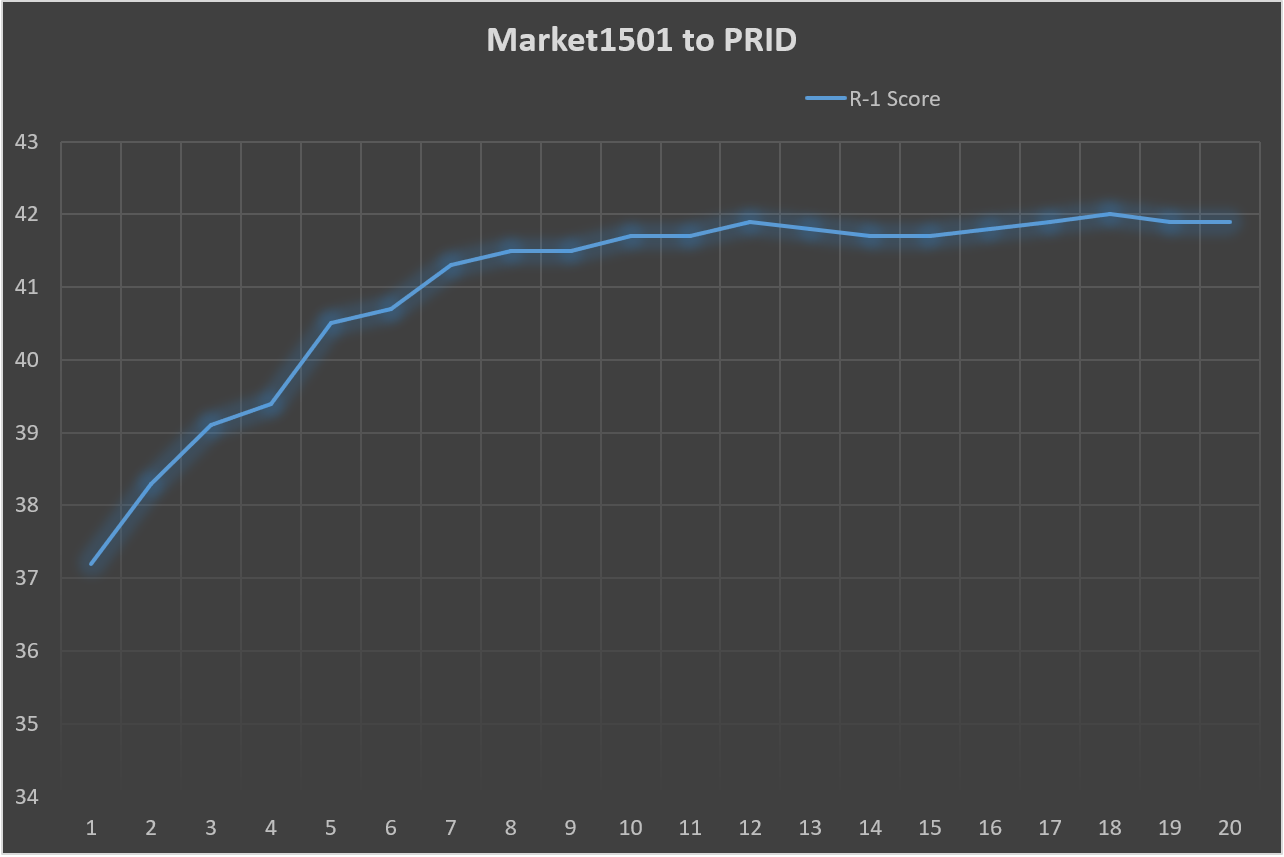}
\caption{Experimental analysis on the parameter M of the multi-modal adaptation for the domain adaptation from Market-1501 to PRID: The horizontal axis denotes the number of M and the vertical axis denotes the R-1 score.
}
\label{im_table}
\end{figure}

\subsection{Ablation Study}
We perform an ablation study to study the contribution of different technical components within our proposed SA-GAN. The ablation study was conducted by using \textit{Duke-MTMC} as the source set and \textit{PRID} as the target sets. Six models are trained as shown in Table \ref{tab:duke2prid} including: 1) `Original' that is trained by \textit{Duke-MTMC} (baseline); 2) SA-GAN(WP) that is trained by using the output of $S_{1}$ without pixel-level adaptation (for validation of the spatial-level adaptation within the proposed SA-GAN); 3) SA-GAN(WD) that uses a normal instead of the proposed disentangle cycle-consistency loss; 4) SA-GAN(WS) that is trained without including the proposed spatial-level adaptation; 5) SA-GAN[M=1] that does not include the proposed multi-modal transformation; and 6) SA-GAN [M=10] that produces 10 adapted images by using 10 random spatial codes.

As Table \ref{tab:duke2prid} shows, `Original' obtains 3.1\% and 2.6\% R-1 scores on PRID-CamA and PRID-CamB due to the large domain shift from \textit{Duke-MTMC} to \textit{PRID}. SA-GAN(WP) and SA-GAN(WS) both outperform the `Original' clearly, demonstrating the contribution of the proposed spatial-level adaptation and pixel-level image adaptation. In addition, SA-GAN[M=1] outperforms both SA-GAN(WP) and SA-GAN(WS) clearly, and this shows that the proposed pixel-level adaptation and spatial-level adaptation are complementary. Further, SA-GAN[M=10] outperform SA-GAN[M=1] clearly, demonstrating the effectiveness of our proposed multi-modal transformation. Note that SA-GAN(WD) performs much worse than SA-GAN[M=1] and this clearly shows the importance of the proposed disentangled cycle loss in concurrent adaptation at pixel and spatial levels.

We provide detailed analysis of the parameter M in Fig. \ref{im_table}. As shown in the figure, the R-1 score keeps rises up with the increasing of M at the ealy stage. When M is lareger than 10, there is minor improvement with increasing of M.

\section{Conclusions}
This paper presents a spatial-aware generative adversarial network (SA-GAN) that achieves domain adaptation at both spatial and pixel levels simultaneously. A multi-modal transformation module is designed which can generate multiple images of different geometric views from one source image. A novel disentangle cycle-consistency loss is designed which can improve both the adaptation stability and adaptation performance clearly. The proposed SA-GAN helps to improve the person ReID performance clearly and it can be easily extended to other tasks.

\bibliographystyle{IEEEtran}
\bibliography{IEEEabrv}

\end{document}